# Challenges for Artificial Cognitive Systems[1]

*Antoni Gomila\* & Vincent C. Müller^*

\*University of the Baleares
^Anatolia College/ACT - University of Oxford

*Abstract:* The declared goal of this paper is to fill this gap: "… cognitive systems research needs questions or challenges that define progress. The challenges are not (yet more) predictions of the future, but a guideline to what are the aims and what would constitute progress." – the quotation being from the project description of EUCogII, the project for the *European Network for Cognitive Systems* within which this formulation of the 'challenges' was originally developed (http://www.eucognition.org). So, we stick out our neck and formulate the challenges for artificial cognitive systems. These challenges are articulated in terms of a definition of what a cognitive system is: a system that learns from experience and uses its acquired knowledge (both declarative and practical) in a flexible manner to achieve its own goals.

[1] This is a final draft of the paper published as: Gomila, Antoni and Müller, Vincent C. (2012), 'Challenges for artificial cognitive systems', *Journal of Cognitive Science,* 13 (4), 453-69. http://doi.org/10.17791/jcs.2012.13.4.453

We want to acknowledge the support of the Comission to the "EuCogII" project in FP7. AG also wants to recognize the support of the Spanish Ministry of Economy and Competitiveness, through project **FFI2009-13416-C02-01**.

**Introduction**

On the grounds of the three intense workshops organized as part of the EuCogII project (Cortona 2009, Rapperswil 2011, Oxford 2012), of the invited talks to the plenary meetings of the network, and of our own research and vision of the field, we have finally produced the present document as an effort to put forward in a clear manner a set of interrelated challenges for artificial cognitive systems, as well as operative ways to measure progress. We have come to the conclusion that a *list of independent challenges* would be senseless, because the potential challenges in such a list would be variously interlinked, in several respects. We have also tried to present the challenges in a *theory- or approach-neutral way*, while at the same time formulated in a way recognizable by the field itself. This is not a naive contention; of course, our understanding of the challenges owes to current theoretical views, so they are clearly not theory-independent. It is just that we have tried not to take sides right from the start among the different research programmes currently active, but made an effort instead of fostering a common ground regarding the problems that all current theoretical approaches have to recognize, as well as about what is to count as progress for any of them.

If this document is to be of any use, then, it must capture the background self-understanding of the field, to drive it in a direction of progress, rather than "imposing" a set of tasks, or appear as a partisan manifest. That's why bibliographical references have been kept to a significant minimum – it is impossible to do justice to such a multidisciplinary field without producing a list of references much longer than the paper itself. The challenges should be acknowledged by everybody, as well as the different strategies available at this moment to tackle them.

In addition, we have tried to formulate them in ways that allow for measurable progress, in a set of well-defined milestones, of increasing achievement. However, our measures of progress do not consist in "competition-like" challenges, where winning is not a guarantee of real progress. It is theoretical progress that it is looked for, rather than technical; hence, progress must be measured in terms of degrees of complexity, or degrees of novelty, or degrees of flexibility – that is, degrees of progress that can only be achieved by theoretical progress. We are aware that this may also sound naive, but we feel confident that if we succeed at the task of characterizing the critical issues in the field, ways to measure progress can follow through as well.

## The task of formulating the challenges and how to approach it

Several possible ways to address our goal are to be avoided, in our view:

a0 As a remake of Hilbert's formalist program for mathematics (or the more recent DARPA 23 challenges for the same discipline); that is, as a list of loosely related problems, so that it is possible to work on one, disregarding the others;

in the case of cognition, an integrated approach is required.

b0 As it is standard in AI-Robotics: in the form of "grand challenges" (like DARPA's driverless vehicles or RoboCup). In these cases, what matters is task success, not advancing in the understanding of how such tasks can be solved;

successful solutions do not necessarily advance our understanding of cognition, because they may rely on "design to the test", i.e. task-specific "tricks".

c0 As an internal agenda for a theoretical approach

> to cognition
>
> the proposed challenges may not be recognized as such by other approaches within the field.

Our goal, accordingly, is to provide a conceptual map of related issues, in a non-partisan way, that can provide orientation regarding what it is already achieved, what's next, how issues relate to one another – and to do so providing milestones, scalable dimensions of progress, which are not bound to be dead alleys, the myopic fine-tuning of strategies that lack generality. To this extent, we need to avoid the assumption that it is just human-inspired, or human-like, artificial systems that matter; even if understanding human cognitive systems may be an outstanding goal, given the central interest in interaction between humans and artificial systems, models and simulations can also target other kinds of natural cognitive systems; but even when the focus is human cognition, there is no need to restrict the goal of the artificial cognitive systems to humanoid robots; morphological resemblance is not required for cognitive interaction between natural and artificial systems. Bio-inspired approaches are welcome, as a way to take advantage of the fruitfulness of collaboration with the cognitive sciences in general, but "human-centrism" is to be avoided as the only approach. Finally, we will try to specify the challenges, not against the best practices/research programs on offer, but taking advantage of them to provide a common plan and vision, a consensus on what should be done first, and what counts as success, given our current understanding of the issues. For this reason, we start with a general characterization of what a cognitive system is, and then proceed next, to articulate the interrelated topics that constitute the challenges were theoretical progress is required in terms of this definition, which provides a clear structure.

## What is a cognitive system – can there be artificial ones?

The way this question is answered, it seems to us, is critical to the specification of challenges. At this basic starting point a critical split can be found in the field, between those researchers who take for granted that, while embodiment is required, cognition still can be thought as computation (Clark, 1998, 2011), on the one hand; and those who inspired by the Artificial Life approach (Steels, 1994) and enactivism (Varela, Thompson & Rosch, 1993), establish a stronger connection between life and cognition, and view cognition as an adaptation (Stewart, Gapenne & di Paolo, 2010).

In order to avoid getting stuck at this starting point, we propose a definition of cognitive system that is not committed to a particular, biological, implementation, and hence, allows for diversity. Our option also departs from the rather frequent attempt to define a rigid hierarchy of orders of complexity for cognitive systems (such as from reactive to deliberative ones, for example, along the vision proposed by (Maynard-Smith & Szathmary, 1997)); this strategy is reminiscent of nineteenth century's view of living being along the "scale of being", whose peak was occupied by "Man". Cognition may be a "major transition" in evolution, but a transition that is characterized by a huge diversity of strategies, of ways of being cognitive; the same applies for artificial systems (Hebb, 2001). What needs to be well defined is what this "cognitive" transition consists in. We submit that a cognitive system is one that learns from individual experience and uses this knowledge in a flexible manner to achieve its goals.

Notice the three elements in the definition: "learning from individual experience", "flexible deployment of such knowledge", and "autonomy" (own goals). A cognitive system is one that it is able to guide its behavior on the knowledge it is able to obtain: it is in this way that it can exhibit flexibility. Systems that come fully equipped with knowledge, or which are not able to gain knowledge out of their experience, or are not able to flexibly use such knowledge in their behavior; or which do not have own goals, are not be counted as cognitive, on this definition. Note that "knowledge" is meant to capture explicit declarative knowledge, but also practical, implicit, knowledge, as well as abilities and skills.

Of course, this is not an innocent or ecumenical notion – as a cursory attention to the debate on 'minimal cognition' would show (van Duij, Keijzer & Franken, 2006). But we find it justified in that it captures the central cases any approach to cognitive systems have to be able to account for. Of course, this definition involves borderline cases, for which it may be difficult (or impossible) to decide whether or not they are "really" cognitive, but this fuzziness is going to appear with any definition. What makes this definition a reasonable one, in our view, is that it definitely focuses on the central cases. In addition, it provides a useful guidance to sort out some of the recurring debates that stem from other definitions on offer. Thus, it avoids considering all living beings as cognitive (reactive, reflex-like systems do not qualify). It also avoids eliminating the possibility of non-living or artificial cognitive beings from the start, which would be question begging. On the other hand, it allows for non-individual learning – or more precisely, it does not rule out evolution as a learning process at the supra-individual level, a learning process that gets expressed in the morphology of the being (Maturana & Varela, 1997; Pfeiffer & Bongard, 2006), but it emphasizes the connection between the learning experience and the flexible use of the knowledge (thus, adaptation per se does not guarantee flexible use of knowledge; therefore, morphology by itself, even if it is the outcome of an evolutionary process of adaptation, does not qualify as knowledge in the relevant sense for cognition). It also leaves space for cultural learning that is transferred to the individual agent in their individual learning experience. On the other hand, it requires more than a syntactical notion of computation for cognition (Fodor, 2010; Anderson, 2003): it requires that what the individual learns is meaningful to the individual, relative to its own goals (di Paolo, 2005).

The different aspects of this definition provide the ground for the representation of the challenges we propose. Advancement is required in how to account for learning from experience; how knowledge is acquired, stored and accessed; how cognitive systems can use it flexibly; and how they can have their own goals; but all of them have to be considered in an integrated way.

**Dealing with an uncertain world**

Natural cognitive beings constitute a way to deal with an uncertain world. This is diametrically opposed to the most common biological strategy: to adapt to just a robust subset of environmental parameters in a rigid manner; or to behave so as to make such parameters rigid or constant; cognitive systems exploit the information available in the environment to adapt in such a way that their behavior is not just dependent upon the current circumstances, but also upon the previous experience. This suggests a relational understanding of world as what's relevant for the system (as the old notion of "Umwelt" proposed by von Uexküll): those parameters that may be relevant to our goals. By learning, cognitive systems try to discover the regularities, constancies, and contingencies that are robust enough to provide such guidance. Learning, though, should not be seen anymore as a passive recording of regularities, as the old empiricism held (Prinz, 2004; Gomila, 2008), but as an active exploration, just like the role of infant active movement is critical in motor development (Gibson, 1979; Thelen & Smith, 1994). In addition, given the relevance of relational contingencies, the materials systems are made of become important.

Talking of an "uncertain world" avoids the ambiguity involved in the alternative notion of "unpredictability", which can be applied both to the world and to the behavior of the system. The notion, though, has to be understood as an epistemic, rather than ontological, one. An uncertain world need not be a noisy, or chaotic, one; just a complex one, that may pose difficulties to a system to anticipate or make sense of what's going on. On the other hand, a cognitive system contributes to the world complexity through its own complex behavior: as long as it behaves in ways that are not predictable just from the specified information about its structure, rules, or inputs, it contributes to the world's complexity.

Measures of progress:
10 Better systems are those able to deal with increasing degrees of world uncertainty -while allowing for increasing environmental variability (in lighting conditions, distances, sizes, time constraints,....).
20 The jump from "virtual" to real environments also counts as progress.
30 In the same vein, systems able to work in different environments (different media: water, air,...; times: day, nights,...; different physical parameters: high and low temperatures, different lighting conditions, pressure, oxigen,...).
40 When moving from "generational" ways of changing the system (as in genetic algorithms) to "tune up" with relevant patterns of information, to individual development of practical skills, of "ways doing things" in a proficient manner, through practice; both take time, but the time-scale is different: generational vs individual.
50 If the system may deal with "social" patterns of information; the idea here is that a social environment is even more complex and demanding.
60 In general, a motivated account of the "initial state" of the system, plus self-organizational development, is required –rather than "ad hoc" assumptions to make the system work; advances in this direction also welcome.
70 Exploration of the properties of new materials, sensors, actuators,... as a way to further explore the environmental "affordances".

**Learning from experience**

This is probably the area where most efforts have been dedicated. There are a multiplicity of techniques and algorithms (broadly, the machine learning area; for an introduction, (Murphy, 2012)) that try to account for this basic cognitive ability. However, these algorithms are generally "information-intensive". Bio-inspired approaches to learning try to find inspiration in the more economical ways natural cognitive systems learn, such as reinforcement learning (Sutton & Barto, 1998), Hebbian learning (Sporns, 2010), dynamic context adaptation (Faubel & Schöner, 2008); while the AI-inspired approaches try to model learning by explicit abstraction (Holyoak, Gentner & Kokinov, 2001). In general, all of these approaches work with abstract data sets, rather than with real environments, and assume a passive view of the system (which is conceived as computational). This seems far from the way natural cognitive systems learn from experience: in an active, situated, way; by exploring the world; and by reconfiguring one's own skills and capabilities. On the other hand, the standard strategy of "annotated" data sets can be seen as a form of social learning, but again passive rather than active.

Measures of progress:
- 80 Systems have to discover environmental patterns, rather than just exhibit forms of non-associative learning, such as habituation or accomodation, which involve some sort of adaptive reaction, but do not provide the system with knowledge to flexibly deploy in behavior.
- 90 Internal states are to function as informational holders of those informational patterns (the effort here is to by-pass a priori debates over representations and how to conceive of them), that can guide behavior. This "common-coding" requirement -that input information is code in a way that can directly guide behavior- relates to Bernstein's problem, the problem of finding the optimal way to move the body/actuators, given a goal (see Bernstein 1967). Because just as there is a problem of combinatorial explosion -the frame problem- for classical AI systems, a risk of combinatorial explosion of degrees of freedom for actuators also appears (Iida, Gómez & Pfeiffer, 2005). The more promising way to block these combinatorial explosion problems is by getting the system configured via its active exploration of the environment (the perception-action loop). This criterion is more difficult to view as scalar, it looks more like an architectural requirement for progress, and it may also involve a further developmental constraint: that the system itself gets build in interaction with the environment.

100    Further, finding structure in one's experience: the recognition of meaningful patterns by recoding, by finding higher-level invariants in these informational patterns, sensorimotor loops (Turvey & Carello, 1986).

110    Finding analogies across domains; that is, relational similarities, rather than just superficial (sensory) similarities (Gentner, 2003).

## How to understand knowledge

Knowledge is the outcome of learning, is what the systems gets when it learns. The current challenge clearly stems from the classical problem of knowledge representation. Classical AI got stuck with the idea of explicit, formal logic-like, propositional representations, and the conception of reasoning as a kind of theorem-proving by transforming those propositional data structures. Together with the aim to formalize expert (or common sense) knowledge, it could not solve the frame problem, the grounding problem, the common-coding problem, etc... New approaches drive attention to practical, embodied, context-dependent, implicit, knowledge skills. But it is not clear yet how this new approach can be carried out (Gomila & Calvo, 2008): how knowledge is codified, implemented, or stored (for how it is accessed, see next section). Success of machine learning methods for classification tasks (via pattern recognition) provide a route to explore, but it has to get more realistic. Another promising approach is brain-inspired dynamical models –which develop the idea that knowledge is in the topology of a network of processing units, plus its coupling to body and environment (Johnson, Spencer & Schöner, 2008). Other approaches are also currently active. In what follows, we try to provide criteria of promising advances that will count in favour of the techniques that are able to achieve them.

Measures of progress:
- 120 Advances from pattern recognition in the input data (data-mining style) to finding relational laws (constancies, affordances) in the environment.
- 130 Advances in multisensory integration, rather than just sensor fusion; sensory-motor contingencies taking different sensors into account, for different environmental dimensions (visual –spatial, auditory –temporal,...), plus proprioceptive information as disambiguating.
- 140 Brain-inspired networks of control (relative to each kind of brain, and each kind of body), for sensory-motor coordination in different tasks: need to go beyond navigation; in particular: a metrics for increasing the repertoire of behaviors available to the system, related to the informational patterns the system is capable to grasp.
- 150 Development, out of this basic, relational, understanding, of a detached, abstract, view of the world (objective knowledge). Psychology teaches that flexible knowledge requires some form of recoding, which is the key to abstraction, to make it adequate to novel, not exactly identical, situations. It can be said the neural networks (specially in their sophisticated forms) account for such abstract recoding, but this is not fully satisfactory, because there is just one network in the model; a different approach is to use layers of neural networks, where the higher level takes as inputs the patterns of the lower, sensory, layers (Sun, 2006), but up to now this is done "by hand". Still another approach, of Vygotskian inspiration, views in the use of public symbols the key to understand cognitive, abstract recoding (Gomila, 2012), but the application of this approach within artificial cognitive systems is just beginning.

**Flexible use of knowledge**

Extracting world regularities and contingencies would be useless unless such knowledge can guide future action in real-time in an uncertain environment. This may require in the end, as anticipated above, behavioral unpredictability, which is a property than runs contrary to the technical requirements of robustness and reliability for artificial systems (to guarantee safety, as the principal engineer's command). The critical issue for flexibility is related to how the knowledge is "stored" (see previous section), and therefore, how it is accessed. The major roadblock to carry this out – regardless of approach – is again combinatorial explosion, whether at the level of propositional representations, as in classical AI, or at the level of degrees of freedom for the control of actuators. But it is also a problem to "judge", in a given situation, which one is the best one to categorize it, given what the system knows.

Different strategies are actively explored as ways to reduce/constrain combinatorial explosion of any kind; it is not possible to establish a clear set of milestones at this point; we would like to suggest the need for exploration of new ideas (different programs may be not incompatible in the end, convergences may emerge).

Measures of progress:
- 160 Simplifying the requirements for cognitive control (context-sensitivity of the "decision" process, distributed adaptive control architectures, potential conflicts adjudicated through accessibility, timing) goes in the right direction.
- 170 Making the controller change in a stochastic way and select the variations that work better (genetic algorithms) has a clear drawback: no individual learning, no flexible deployment of knowledge; progress requires the integration of interaction with the environment, the learning process and the knowledge acquisition and its flexible deployment.
- 180 Dynamicist approaches (such as the "dynamic field approach", (Johnson, Spencer & Schöner, 2008)), which look for system criticalities in the state space, and resort to force field metaphors of distributed activation, hold a potential to advance on a control architecture which is not homuncular, which is not committed to a "central executive", whose powers remain mysterious: control emerges out of a distributed field of activation.
- 190 Exploiting the body (sensors and actuators), and their dynamic loops, is also a promising way to constraint the explosion of combinatorial options, both at the representational and behavioral control ends (Philipona, O'Regan, & Nadal, 2003).
- 200 Advances in "schematization" of sensory-motor contingencies, as a simple way of recoding (in fact, a form of abstraction) also count as progress.
- 210 The use of heuristics ("fast and frugal", non foul-

           proof, context-sensitive, procedures; Gigerenzer, Hertwig, & Pachur, 2011), instead of algorithms, also provides a way to avoid combinatorial explosion of algorithmic processes, but it requires some form of adjudication, to determine which one gets in charge (again, the activation metaphor provides a way to deal with this problem).

220      Emotions can also provide a path for progress, in their role as quick valuations or assessment of situations, on simple hints, from the point of view of the system. Brain-inspired models of the reward system, of the amygdala; reinforcement-based expected reward, rather than calculation of expected utility, are strategies which could make progress in this regard (Sutton & Barto,1997).

**Autonomy**

Autonomy is related to agency, and agency to having own goals. It requires internal motivation, and a sense of value "for the system". It also requires some kind of "self-monitoring": an internal grasp of one's cognitive activity is required to make possible the "internal error detection" (Bickhardt, 2008), as the central cognitive capacity of self-monitoring – involving both whether the behavior matches the relevant intention, and whether it is carried out as intended.

   In systems like us, this property is achieved by a double control architecture: the autonomous nervous system (including the endocrine one), plus the central nervous system; both systems are also interrelated. In general, a cognitive system involves a basic regulatory system that implicitly defines the needs and requirements, the motivations and homeostatic goals of the system, for which internal sensory feedback is required to keep the system within the range of vital parameters. In addition, a central system allows for more sophisticated forms of environmental coupling, for informational management, for memory and learning, and for control contingent on such previous experience. A full-blown agent, from this point of view, is one, which is capable to generate new behavior appropiate to new circumstances (which seems impredictible just given the situation); it requires self-organization, a homeostatic relationship with the environment of self-sustained processes (di Paolo, 2005; Moreno & Etxebarria, 2006) -something still very far from current technology. It may also require the ability to "work off-line", to recombine previous experiences, and to test in the imagination the new options (Grush, 2004). Autonomy comes in degrees and it is a necessary feature of systems that can deal with the real world (Müller 2012).

Measures of progress:
- 230 When programming of all possibilities (look and search strategy in the problem space) is no longer the basic strategy; progress requires to let the artificial system "go beyond" the programmed, to modify itself, to choose among several options, to choose which knowledge to use, … according to the contingencies of its individual experience.
- 240 From systems with externally imposed goals in a non-previously specified manner in a previously specified environment (simplified, virtual); to similar systems able to deal with non-previously specified environments; from systems that can choose among several pre-specified goals according to circumstances, to systems that can develop new goals; from systems that can modify/change themselves according to circumstances to systems able to solve internal motivational conflicts (change goals).
- 250 From "simple" systems, whose behavior depend upon a few parameters, to more "complex" ones –by

increasing the number of parameters, and letting them to interact non-linearly, complexity follows.

**Social cognitive systems**

Social cognitive systems address this learning process in a facilitated way, by starting in a simplified, structured environment; by receiving feedback and scaffolding from others; by using others as models (Steels, 2011).

Of course, this creates a specific problem of social learning: to find out in the first place which parts of one's world are other cognitive systems, and to discover the regularities, constancies, and contingencies, that are relevant in this area. All this is especially relevant for the area of interaction among cognitive beings, both natural and artificial.

It has also become clear that increasing autonomy in the interaction between natural and artificial systems requires some kind of "moral control"; the attempt to guarantee that the interaction doesn't turn against the human (Arkin, 2007; Wallach & Allen, 2009).

Measures of progress:
- 260  Systems able to interact with other systems in increasingly complex ways -from simple synchronization, to imitation, to emulation, to cooperation, to joint action (Knoblich & Sebald, 2008).
- 270  Systems able to develop "common worlds", a common understanding of how things go (share knowledge, distribute tasks according to abilities, …)
- 280  Increasing "mental" abilities – which cannot be dissociated from artificial systems abilities (to recognize "sadness" or "rage" in a human the artificial system is required to be able to "express" emotional states (Vallverdú & Casacuberta, 2009; progress in this area involves not just facial mimicry, but understanding of

> emotional states as expressed, and simulation of such expression process.
> 290    In fact, this emotional capacity constitutes the ground level of our moral understanding (Damasio, 1997; Greene et al., 2001), and this suggest that progress in this regard involves modelling moral emotions (Gomila & Amengual, 2009).
> 300    Proficiency in pragmatic linguistic interaction in naturalistic contexts, as a way to interact, coordinate and cooperate (Tikhanoff, Cangelosi & Metta, 2011).

## 10. Conclusion

As intended progress in one challenge is not independent on progress on many others – the typical property of cognition is an integration of capabilities and elements. It is not possible, though, to establish milestones at this global level, because of the intrinsic diversity of cognitive beings. What it does seem advisable at this point is to emphasize integrated systems over specialized algorithms. Classical AI has worked under the assumption of modularity, as engineering in general: the goal is to add new facilities to a system without having to re-design it anew. There is reason to doubt this assumption is going to work for cognitive systems – the scaling problem is a serious one. New capabilities may require some sort of reorganization, in non-principled ways. Hence, a final, global, challenge, concerns this problem of scaling-up cognitive systems –which may induce a vision of the field of artificial cognitive systems itself as following an evolutionary trajectory, hopefully one of increasing fitness.